\documentclass{article}

\usepackage[preprint]{neurips_2021}

\usepackage[utf8]{inputenc} 
\usepackage[T1]{fontenc}    
\usepackage{hyperref}       
\usepackage{url}            
\usepackage{booktabs}       
\usepackage{amsfonts}       
\usepackage{nicefrac}       
\usepackage{microtype}      
\usepackage{xcolor}         

\usepackage{graphicx}
\usepackage{amsmath}
\usepackage{amsthm}
\usepackage{amssymb}
\usepackage{cleveref}
\usepackage{booktabs}
\usepackage{subfigure}
\usepackage{algorithmic}
\usepackage{algorithm}

\newcommand{\IE}{\textit{i.e.}, }
\newcommand{\EG}{\textit{e.g.}, }

\newcommand{\ROID}{\textit{DOI}}

\newcommand{\E}{\operatorname{\mathbb{E}}}
\newcommand{\pin}{p_{in}}
\newcommand{\pout}{p_{out}}
\newcommand{\pmix}{p_{mix}}

\title{DOI: Divergence-based Out-of-Distribution Indicators via Deep Generative Models}

\author{
Wenxiao Chen, Tsinghua University \And Xiaohui Xie, Tsinghua University \And Mingliang Li, China Unicom \And Dan Pei, Tsinghua University
}

\begin{document}

\maketitle

\begin{abstract}
To ensure robust and reliable classification results, OoD (out-of-distribution) indicators based on deep generative models are proposed recently and are shown to work well on small datasets. In this paper, we conduct the first large collection of benchmarks (containing 92 dataset pairs, which is 1 order of magnitude larger than previous ones) for existing OoD indicators and observe that none perform well. We thus advocate that a large collection of benchmarks is mandatory for evaluating OoD indicators. We propose a novel theoretical framework, DOI, for divergence-based Out-of-Distribution indicators (instead of traditional likelihood-based) in deep generative models. Following this framework, we further propose a simple and effective OoD detection algorithm: Single-shot Fine-tune. It significantly outperforms past works by 5$\sim$8 in AUROC, and its performance is close to optimal. 
In recent, the likelihood criterion is shown to be ineffective in detecting OoD. Single-shot Fine-tune proposes a novel fine-tune criterion to detect OoD, by whether the likelihood of the testing sample is improved after fine-tuning a well-trained model on it. Fine-tune criterion is a clear and easy-following criterion, which will lead the OoD domain into a new stage.
\end{abstract}

\section{Introduction}
\footnote{This paper is developed at the same time as \cite{zhisheng2020likelihood} independently. 
The key idea of this paper is very similar to \cite{zhisheng2020likelihood}. This paper only has incremental content (theorems, methods, and experiments) compared with \cite{zhisheng2020likelihood}. 
Since \cite{zhisheng2020likelihood} has been published in NIPS 2020, this paper is provided here for the community to share our large-scale experiments and theorems. The code is in \url{https://github.com/chenwenxiao/DOI}.
}
Machine learning has achieved impressive success in the classification domain through deep neural network classifiers~\cite{szegedy2016inception,he2016deep,zagoruyko2016wide}. Knowing when a machine learning (ML) model is qualified to make predictions on input is critical to the safe deployment of ML technology in the real world~\cite{choi2018waic}. When training distribution (called in-distribution) differs from testing distribution (called out-of-Distribution), neural networks may provide (with high confidence) arbitrary predictions on inputs that they are unaccustomed to seeing. This is known as the Out-of-Distribution (OoD) problem\cite{choi2018waic}.
%
For example, a classifier trained on CIFAR-10~\cite{krizhevsky2009learning} may recognize the house number in SVHN~\cite{netzer2011reading} as a horse, which might lead to potential risks.

Therefore, it is crucial to develop \textbf{OoD indicators} for detecting whether a testing sample is from in-distribution or out-of-distribution to ensure that applications based on classifiers are robust and reliable. The common belief~\cite{bishop1994novelty} is that the OoD indicators can be based on density model: train a density model $p_\theta(x)$ (as an OoD indicator) to approximate the empirical distribution of training data, and refuse the sample $x$ when $p_\theta(x)$ is sufficiently low.
However, recent works~\cite{nalisnick2019do,choi2018waic,hendrycks2018deep} show that density estimates by \textit{deep generative models}~\cite{dinh2016density,tomczak2018vae,takahashi2019variational,van2016conditional}, which generate realistic samples, assign higher density to samples from out-of-distribution. For example, according to ~\cite{nalisnick2019do}, this phenomenon occurs in CIFAR-10 (as in-distribution) vs SVHN (as out-of-distribution) for different likelihood-based models, while the data in CIFAR-10 and SVHN have significant different semantics.

To alleviate the aforementioned phenomenon, more advanced OoD indicators \cite{serra2019input,song2017pixeldefend,choi2018waic,ren2019likelihood,song2019unsupervised,che2019deep} are proposed recently based on deep generative models and shown to perform well on few datasets,\EG the number of dataset pairs (in-distribution dataset, out-of-distribution dataset) is only $1$ to $10$.
However, a \textit{robust} OoD indicator should detect samples from any out-of-distribution~\cite{chen2020robust}. Thus it should be evaluated on a large collection of benchmarks. In this paper, we first conduct a large collection of benchmarks with 92 dataset pairs (based on 14 popular image datasets, including MNIST, FASHION-MNIST, KMNIST, NOT-MNIST, Omniglot, CIFAR-10, CIFAR-100, TinyImagenet, SVHN, iSUN, CelebA, LSUN, Noise and Constant), whose scale is 1 order of magnitude larger than all above works. We observe that none of the above OoD indicators perform well on the large collection of benchmarks (see later in Table~\ref{tab1}). Based on this observation, we advocate that \textit{experiments on few datasets are unreliable, and a large collection of benchmarks is mandatory for evaluating OoD indicators}.

Another interesting observation that we discover by accident, as a result of the fact that we try to enumerate the dataset pairs when possible (\EG (CIFAR-10, SVHN) and (SVHN,CIFAR-10) are both in our experiment setting), is that, on all dataset pairs, \textit{$p_\theta(x) > p_\omega(x)$ when $x$ is from in-distribution and $p_\theta(x) < p_\omega(x)$ when $x$ is from out-distribution}, where $\omega$ is a generative model trained on the out-of-distribution data.
Inspired by this intuitive (in retrospect) observation, we propose a fundamental theoretical framework~\ROID{} for \textbf{D}ivergence-based \textbf{O}ut-of-Distribution \textbf{I}ndictors in deep generative models. Following this framework, we further propose a simple and effective out-of-distribution detection algorithm, Single-shot Fine-tune algorithm with three mainstream deep generative models (VAE, PixelCNN, and RNVP).
In our experiments, Single-shot Fine-tune significantly outperforms existing works by 5$\sim$8\% in AUROC, and its performance is close to the theoretical optimal results.

The main contributions of this paper are as follows:
\begin{itemize}
\item We conduct the first large collection of benchmarks (containing 1 order of magnitude larger datasets than previous ones) for existing OoD indicators and observe that none perform well. We thus advocate that \textit{experiments on few datasets are unreliable, and a large collection of benchmarks is mandatory for evaluating OoD indicators}.
\item We propose a novel theoretical framework~\ROID{} for divergence-based out-of-distribution indicators (instead of traditional likelihood-based) in deep generative models. Following \ROID{}, we further propose a simple and effective algorithm,  Single-shot Fine-tune. We believe that \ROID{} framework could guide the development in the OoD domain.
\item {Single-shot Fine-tune algorithm significantly outperforms past works by 5$\sim$8\% in AUROC. Single-shot Fine-tune is the first fine-tuned-based inductive OoD method, which shows that fine-tune criterion is a practical and effective criterion for detecting OoD. In fine-tune criterion, the use of both knowledge and cognition improves the performance significantly, which will draw attention to the cognition ability of deep generative models.} 
\end{itemize}

\section{Background}
Likelihood-based generative models are widely viewed to be robust to detect out-of-distribution samples by the model density intuitively. However, the densities of common likelihood-based models, \EG RealNVP~\cite{dinh2016density}, VAE~\cite{tomczak2018vae,takahashi2019variational} and PixelCNN~\cite{van2016conditional}, have been shown to be problematic for detecting out-of-distribution data~\cite{nalisnick2019do}. These likelihood-based models assign a higher likelihood for samples from SVHN (out-of-distribution) than samples from CIFAR-10 (in-distribution).

To solve this problem, some researchers proposed some variants of these models for detecting out-of-distribution data~\cite{che2019deep} and some researchers proposed improved indicators to replace log-likelihood on common models~\cite{serra2019input}. Common models are widely applied in the images domain, but variants are not. Moreover, evaluating numerous variants on the large collection of benchmarks is more expensive, while common models are easy to train, and many indicators can share one well-trained model. Furthermore, it is necessary to check the generality of indicators on common models. By the above motivations, this paper focuses on the indicators based on common models.

\cite{song2017pixeldefend} proposed permutation tests statistics $T_{perm}(x)$ as the indicator to detect OoD. The rank of $p_\theta(x)$ in the training set is used as OoD indicators. Both low-likelihood and high-likelihood samples are identified as OoD. It is significantly useful to solve the counterexample of CIFAR-10 vs SVHN~\cite{nalisnick2019do}.

\cite{choi2018waic} used Watanabe Akaike Information Criterion (WAIC) based on model ensembles.
\begin{equation}
	\text{WAIC}(x) = \mathbb{E}_{\theta} [\log p_\theta(x)] - \operatorname{Var}_{\theta} [\log p_\theta(x)]
\end{equation}

\cite{ren2019likelihood} proposed a likelihood ratio indicator for deep generative models. They proposed a background model $p_{\theta_0}(x)$ to capture the general background statistics and a likelihood ratio indicator $LLR(x)$ to capture the significance of semantics compared to the background model.
\begin{equation}
LLR(x) = \log p_\theta(x) - \log p_{\theta_0}(x)
\end{equation}

\cite{serra2019input} observed that input complexity excessively affects the generative models' likelihoods. Then an estimation is proposed for input complexity $L(x)$, to derive a parameter-free OoD indicator $S(x)$:
\begin{equation}
	S(x) = -\log p_\theta(x) - L(x)
\end{equation}

\cite{song2019unsupervised} observed that generative models with batch normalization assign a lower likelihood to OoD samples than in-distribution samples.
Meanwhile, the corresponding log-likelihood decreases dramatically for OoD samples, but is relatively stable for in-distribution samples, as the ratio of test samples in a batch increases.
Based on the insight, $T_{b, r_1, r_2}(x)$, measuring the difference of log-likelihood under two situation that ratio of test samples are different, is proposed for OoD detection.

Some researchers also proposed to use labels (for classification tasks) to solve OoD.
\cite{che2019deep} proposed $p(x|y)$ for OoD detection. It uses conditional deep generative models to verify the predictions of classifier. \cite{alemi2018uncertainty} use VIB to model the bottleneck $I(Z;Y)-\beta I(Z;X)$ where $I$ is the mutual information.
\cite{hendrycks2016baseline,hendrycks2018deep,hsu2020generalized,lee2018simple,lakshminarayanan2017simple} proposed some indictors based on classifier for detecting OoD.

\section{Problem Statement}\label{sec2}

Out-of-distribution detection problem can be formulated as a special binary classification problem. In canonical binary classification problem, a set of images with label 1 (denoted by $\mathcal{D}^{train}_{in}$) and a set of images with label 0 (denoted by $\mathcal{D}^{train}_{out}$) are given in training; in testing, a set of images without label (denoted by $\mathcal{D}^{test}$) are given, algorithm needs to predict the label of each image in $\mathcal{D}^{test}$.  $\mathcal{D}^{test}$ consists of $\mathcal{D}^{test}_{in}$ and $\mathcal{D}^{test}_{out}$.  $\mathcal{D}^{train}_{in}$ and  $\mathcal{D}^{test}_{in}$ are sampled from   dataset $D_{in}$ and $\mathcal{D}^{train}_{out}$ and  $\mathcal{D}^{test}_{out}$ are sampled from another dataset $D_{out}$.

It is the key difference between OoD detection and canonical binary classification that \textbf{$\mathcal{D}^{train}_{out}$ is unkown in OoD detection problem}. Moreover, $\mathcal{D}_{out}$ is quite distinct from $\mathcal{D}_{in}$ in OoD problem, \EG $\mathcal{D}_{in}$ are animals and $\mathcal{D}_{out}$ are house numbers.
$\pin$ and $\pout$ denote corresponding data distributions of $\mathcal{D}_{in}$ and $\mathcal{D}_{out}$, where $\pin$ is called in-distribution and $\pout$ is called out-of-distribution.

It is important to decide whether two datasets are distinct on the large collection of benchmarks. Two datasets $\mathcal{D}_{in}$ and $\mathcal{D}_{out}$ are called \textbf{simply-classified} if a common classifier (\EG ResNet34) trained for 2-class classification, when $\mathcal{D}^{train}_{in}$ and $\mathcal{D}^{train}_{out}$ are both known, can simply predict the accurate label for images in $\mathcal{D}^{test}$ (AUROC $\geq 99.9\%$). If two datasets A, B are simply-classified, A vs B and B vs A dataset pairs will be considered in our experiments.

This paper considers the OoD detection based on common deep generative models, \IE VAE, PixelCNN, flow-based models, and GANs. We focus on searching for a simple and effective indicator for OoD. More common datasets (shown in Section~\ref{sec6}) are used to validate the generality of indicators.

All common metrics, including AUROC, AUPR, AP, FPR@TPR95, are considered in this paper. AUROC is selected as the major metric, and other metrics are shown in appendix B. AUROC is a threshold-independent metric~\cite{davis2006relationship} and is widely used in the OoD domain.  

\section{Motivating Observations}\label{sec3}

\begin{figure}[t]
\centering
\subfigure[CIFAR-10 vs SVHN]{
\includegraphics[width=0.22\columnwidth]{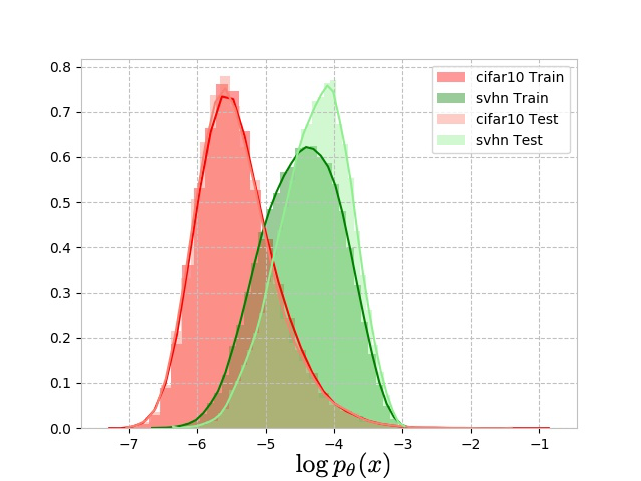}
}
\subfigure[KMNIST vs Omniglot]{
\includegraphics[width=0.22\columnwidth]{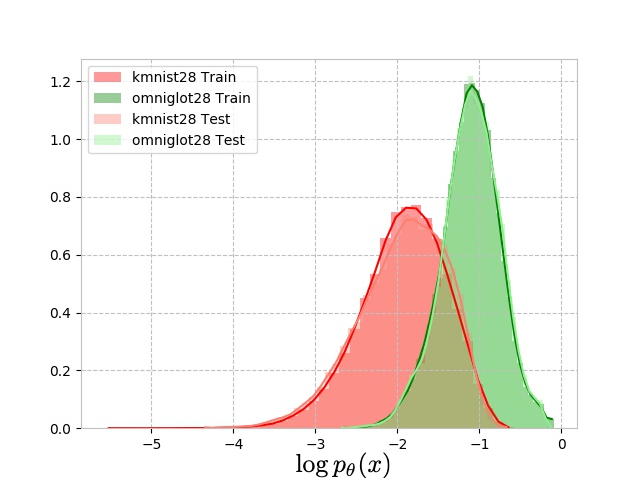}
}
\subfigure[SVHN vs CIFAR-10]{
\includegraphics[width=0.22\columnwidth]{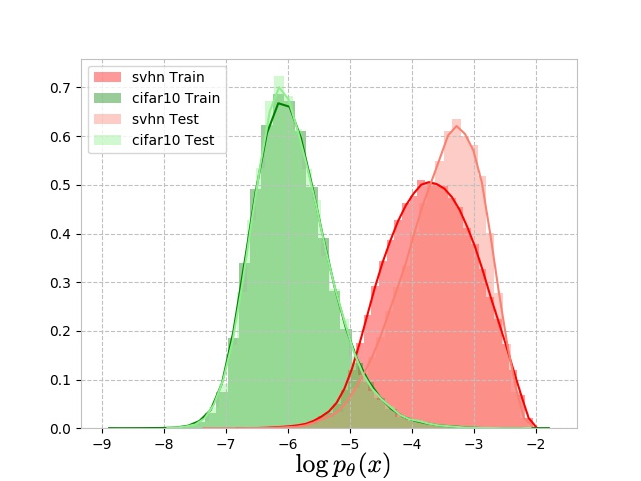}
}
\subfigure[MNIST vs Omniglot]{
\includegraphics[width=0.22\columnwidth]{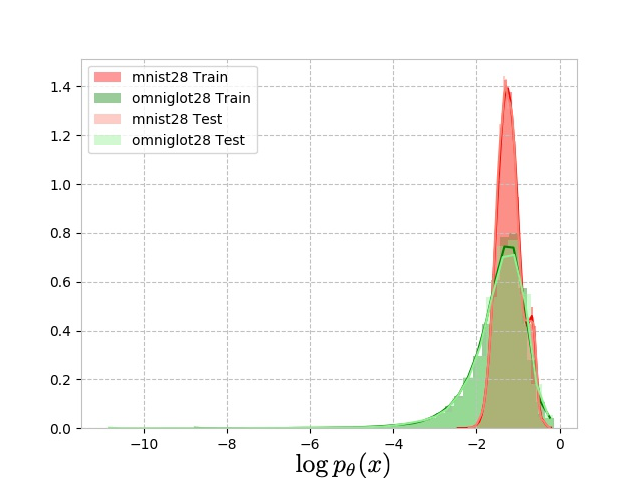}
}
\caption{The histogram of log-likelihood of VAE.
The green and red parts denote the log-likelihood of out-of-distribution and in-distribution, respectively.
Intuitively, the log-likelihood of out-of-distribution is expected to be higher than in-distribution.
However, in the above experiments, the likelihood of out-of-distribution might be higher, lower, or nearly the same as in-distribution.
In above figures, the AUROC of log-likelihood is 8\%, 9\%, 99\%, 59\% and the AUROC of $T_{perm}$ is 84\%, 82\%, 98\%, 66\%. }
\label{fig1}
\end{figure}

\subsection{Counterexamples}
\label{subsec:counterexamples}
Intuitively, the log-likelihood of the sample from out-of-distribution is expected to be lower than the in-distribution because models are trained on in-distribution.
However, \cite{nalisnick2019do} observes that VAE, PixelCNN, and RealNVP all assign the higher log-likelihood to samples from out-of-distribution in experiments CIFAR-10 vs SVHN and NotMNIST vs MNIST.
The number of datasets in \cite{nalisnick2019do} is quite small, and we suspect that there are more counterexamples on the large collection of benchmarks.

Therefore, we reproduce the experiments on a large collection of benchmarks and find 28 counterexamples in 92 dataset pairs, as shown in Figure~\ref{fig1} and appendix A.
These experiments show that log-likelihood is unpredictable at out-of-distribution, \IE it might be lower, higher, or same to in-distribution.
Moreover, the methods based on the log-likelihood might have counterexamples on the large collection of benchmarks.
We reproduce the indicators~\cite{alemi2018uncertainty,song2017pixeldefend,ren2019likelihood,song2019generative,nalisnick2019do,che2019deep,alemi2018uncertainty} on common generative models and find counterexamples, shown in appendix A. 
Especially, \cite{nalisnick2019do} observed that there is a clear correlation between likelihoods and complexity estimates. We checked their observation on the large collection of benchmarks, shown in Figure~\ref{fig2}. 

Furthermore, counterexamples for OoD indicators not based on deep generative models are shown in appendix A. \EG \cite{lee2018simple} reaches 98.24\% AUROC on SVHN vs CIFAR-10, but only 38.22\% AUROC on Omniglot vs FashionMNIST. These counterexamples indicate the critical generality problem in the OoD domain. An important reason for this phenomenon is that \textit{OoD indicators are always designed based on motivating observations only on few datasets. However, it is not guaranteed that these observations are also established on the large collection of benchmarks}. These counterexamples encourage the evaluation on the large collection of benchmarks.

\begin{figure}[t]
\centering
\subfigure[Correlation]{
\includegraphics[width=0.22\columnwidth]{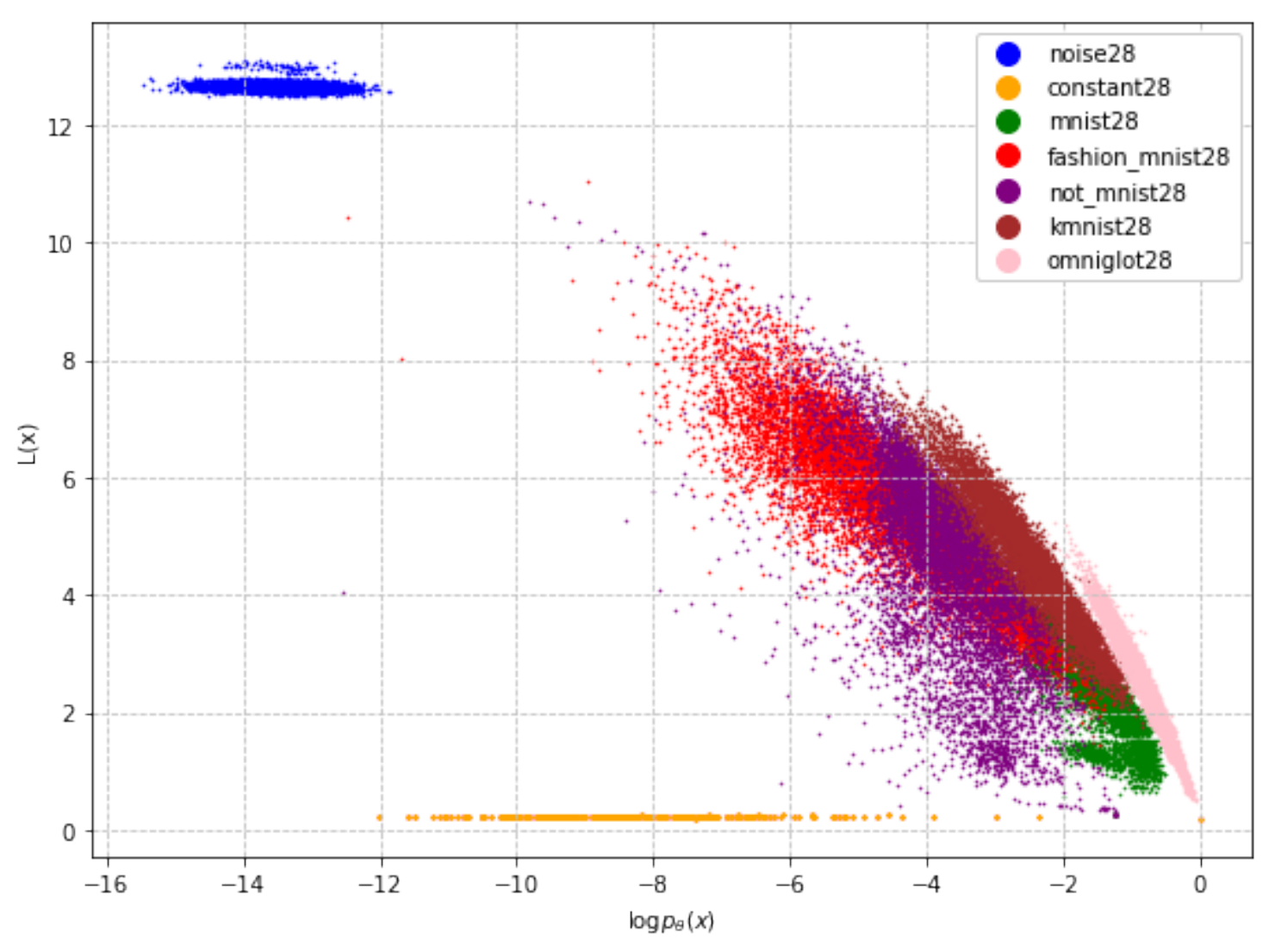}
}
\subfigure[AUROC = 0.9867]{
\includegraphics[width=0.22\columnwidth]{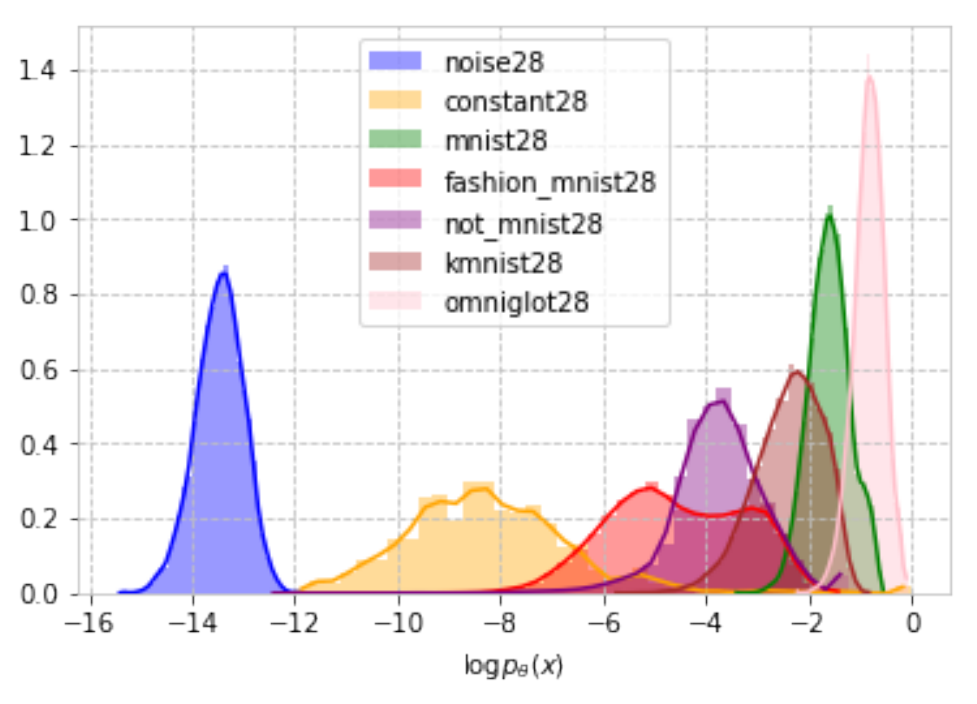}
}
\subfigure[AUROC = 0.7770]{
\includegraphics[width=0.22\columnwidth]{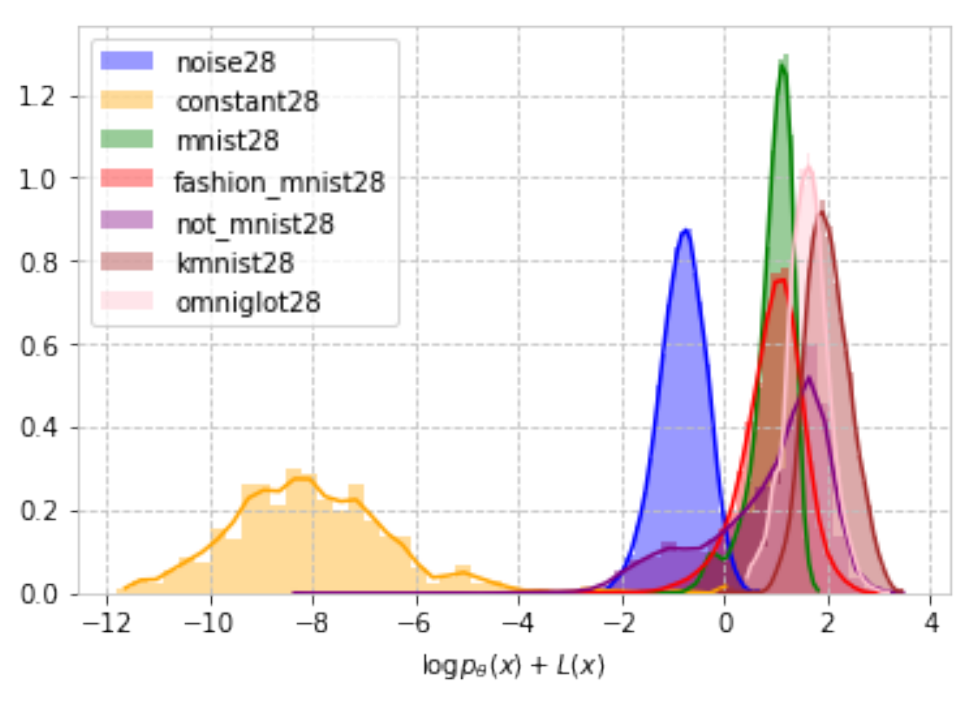}
}
\subfigure[AUROC = 0.9999]{
\includegraphics[width=0.22\columnwidth]{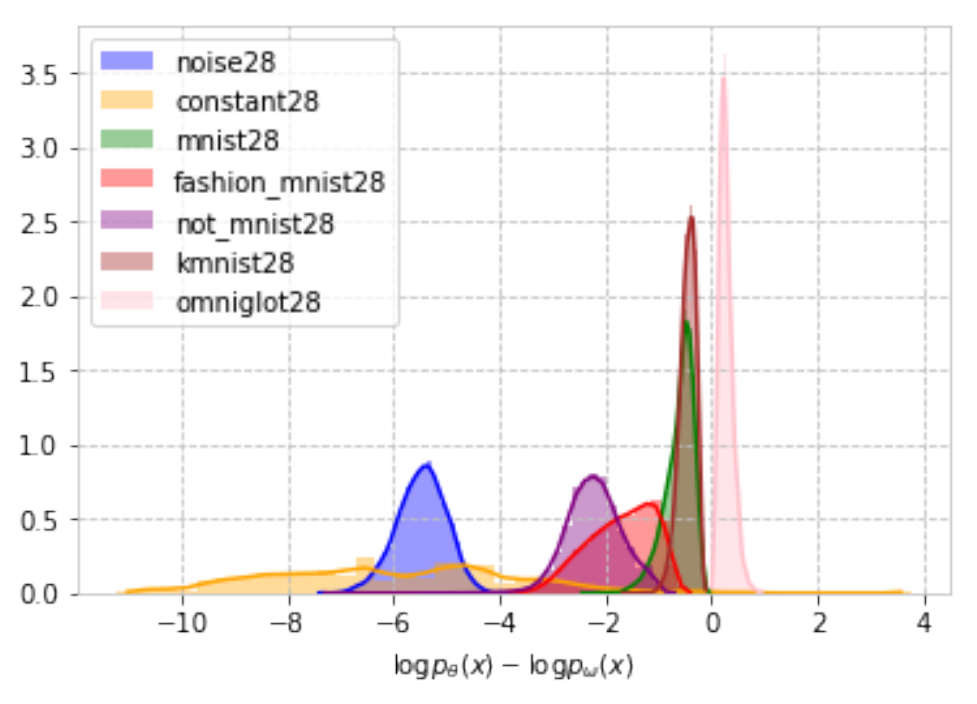}
}
\caption{Omniglot is in-distribution, and other datasets are out-of-distribution. (a) shows the correlation between likelihoods trained on Omniglot and complexity estimate. (b) shows the histogram of log-likelihood. (c) shows that indicator $S(x)$ might perform worse than log-likelihood. It means that $L(x)$ is rough, and we need a more precise, stable, and interpretable estimate to assist log-likelihood for detecting OoD. (d) shows $\log \frac{p_\theta(x)}{p_\omega(x)}$ might be a good choice as a theoretical indicator (NOT practical indicator) where $p_\theta$ is trained on $\mathcal{D}^{train}_{in}$ and $p_\omega$ is trained on $\mathcal{D}^{train}_{out}$.
}
\label{fig2}
\end{figure}

\subsection{Performance on large collection of benchmarks}

For the following reasons, a large collection of benchmarks is used:

\noindent \textbf{1) Check observations} OoD indicators of past works are based on the motivating observation on few datasets. However, we find that the observation on few datasets is not reliable on the large collection of benchmarks, as shown in Section~\ref{subsec:counterexamples}. In practice, OoD indicators need to handle arbitrary images, and it will be harmful if  OoD indicators can only work for few datasets. Therefore, it is necessary to validate the generality of motivating observation on the large collection of benchmarks.



\noindent \textbf{2) Average Performance} Average performance on the large collection of benchmarks is better for assessing indicators. In CelebA vs LSUN, log-likelihood reaches 98\% AUROC. However, it only reaches 2\% in CelebA vs SVHN in appendix A. The average performance will consider such experiments with lower AUROC. It is more meaningful to improve the average performance of indicators than to improve little (\EG 99.1\% to 99.2\%) in a single experiment. 

Indicators of previous works via common deep generative models do not perform well on the large collection of benchmarks, as shown in Table~\ref{tab1}, where
DeConf-C, MCMC Recon, MCMC $\log p_\theta(x)$, $D_\theta(x)$, $\|\nabla_x D_\theta(x)\|$, entropy, $\max_{y} p(y|x)$, Mahalanobis, ODIN and disagreement are proposed by past wroks~\cite{hendrycks2016baseline,hsu2020generalized,lee2018simple,alemi2018uncertainty,liang2018enhancing,kumar2019maximum,xu2018unsupervised,chen2019unsupervised,lakshminarayanan2017simple}.
Thanks to the assistance of $L(x)$, $S(x)$ performs well among past works, which encourages us to develop better assistance.

\begin{table}[t]
\caption{The average AUROC of past works and our method Single-shot Fine-tune (SF(x)) with VAE, PixelCNN, and RNVP on the large collection of benchmarks.
SF(x) outperforms past works.
}
\label{tab1}
\begin{minipage}{0.55\textwidth}
\begin{tabular}{lllllll}
Indicator     & VAE & PixelCNN & RNVP  \\
\toprule
$\log p_\theta(x)$ & 70.11 & 72.26 & 69.19 \\
$T_{perm}(x)$ & 89.71 & 84.28 & 89.72 \\
$\|\nabla_x \log p_\theta(x)\|$ & 53.95 & NA & 24.27 \\
$LLR(x)$ & 69.47 & 77.46 & 64.03\\
$WAIC(x)$ & 74.59 & 82.19 & 83.74 \\
$\mathop{Var}_\theta [\log p_\theta(x)]$ & 83.11 & 82.21 & 86.06 \\
$\log p(x|y)$ & 53.13 & 69.22 & 71.27\\
$T_{b, r_1, r_2}(x)$ & 67.26 & 56.98 & 77.38 \\
BN $\log p_\theta(x)$ & 85.15 & 64.10 & 82.00 \\
$S(x)$ & 81.88 & 88.33 & 80.11\\
SF(x) & \textbf{95.78} & \textbf{97.64} & \textbf{94.34} \\
\bottomrule
\end{tabular}
\end{minipage}
\begin{minipage}{0.4\textwidth}
\begin{tabular}{lllllll}
Indicator     & Model & AUROC  \\
\toprule
Recon & VAE & 69.26 \\
ELBO & VAE & 69.44  \\
ELBO - Recon & VAE & 52.85 \\
MCMC Recon & VAE & 67.43  \\
MCMC $\log p_\theta(x)$ & VAE & 67.45 \\
Volume & RNVP & 62.46  \\
$\log p_\theta(z)$ & RNVP & 74.58 \\
H & VIB & 66.79 \\
R & VIB & 58.78 \\
$D_\theta(x)$ & WGAN & 79.15  \\
$\|\nabla_x D_\theta(x)\|$ & WGAN & 60.55 \\
Disagreement & ResNet & 69.25 \\
Mahalanobis & ResNet & 83.02 \\
Entropy of $p(y|x)$ & ResNet & 62.74 \\
$\max_{y} p(y|x)$ & ResNet & 61.59 \\
ODIN & ResNet & 60.68 \\
DeConf-C & ResNet & 68.59 \\
DeConf-C* & ResNet & 71.09  \\
\bottomrule
\end{tabular}
\end{minipage}
\end{table}

\subsection{Observation of KL-based indicator}
As shown in Figure~\ref{fig2}, the complexity estimate is unstable, and sometimes it might lower the performance. From the view of complexity estimate, $L(x) = -\log_2 p(x|\mathcal{M}_0)$ is the log-likelihood of a universal model $\mathcal{M}_0$~\cite{serra2019input}. Therefore, we try to find another likelihood-based term to replace $L(x)$.
In experiments on large collection of benchmarks, we observe a common phenomenon (for 99.815\% data in all experiments) that $p_\theta(x) < p_\omega(x)$ for most $x$ in $D^{test}_{out}$ and $p_\theta(x) > p_\omega(x)$ for most $x$ in $D^{train}_{out}$, where $p_\omega(x)$ is a likelihood-based model trained on $D^{train}_{out}$.
The average AUROC of $\log p_\theta(x) - \log p_\omega(x)$ reaches nearly 100\% on all datasets in Table~\ref{tab3}.
$\log p_\omega(x)$ can be seen as an improvement of $L(x)$, which is not a universal model but a particular model for OoD detection. $\log p_\theta(x) - \log p_\omega(x)$ is called \textbf{KL-based indicator} .

However, $D^{train}_{out}$ is unknown in the OoD problem. Therefore, the KL-based indicator is only a theoretical indicator, not practical. Next, we develop an indicator approximating to KL-based indicator without training on $D^{train}_{out}$ and explain why KL-based indicator is always effective theoretically.

{
\section{Algorithm}\label{sec5}
Based on the observation of the KL-based indicator, we propose a novel theoretical framework \ROID{}. Through \ROID{}, a strawman algorithm, Naive Fine-tune, is proposed for introducing a novel OoD criterion, called fine-tune criterion. Naive Fine-tune has an obvious weakness that it needs $D^{test}$, which is not allowed in the OoD domain.
To solve this problem, we propose Single-shot Fine-tune algorithm, which fine-tunes the model on the single testing sample. It is an inductive method.

\subsection{Divergence-based OoD Indicators}
We propose a fundamental theoretical framework for \textbf{D}ivergence-based \textbf{O}oD \textbf{I}ndicators called \ROID{}. 
\textbf{The key idea of \ROID{} is to use the divergence between in-distribution and out-of-distribution to detect OoD instead of likelihood. } To achieve this idea, 
Kullback-Leibler divergence is chosen as the divergence in \ROID{}. Based on 5 fundamental assumptions (also observed in experiments), many theorems are proved in appendix C. We show the theorems without proves here. 

\textbf{Theorem. 1} \textit{
	$\log \pin(x) - \log \pout(x)$ and $\log p_\theta(x) - \log p_\omega(x)$ and are effective symmetric indicators, \IE the two indicators both reach same performance in experiment A vs B and B vs A, with threshold 0, where $p_\theta \rightarrow \pin$ and $p_\omega \rightarrow \pout$. $\log p_\theta(x)$ maps $\pin$ into a gaussian distribution. }
	
\textbf{Theorem. 2} \textit{
	For any mixture distribution $\pmix = \alpha \pin + \beta \pout$ where $\alpha + \beta = 1$ and $\alpha, \beta > 0$, the performance of indicator $\log \pin(x) - \log \pmix(x)$ and indicator $\log \pin(x) - \log \pout(x)$ is equal for OoD detection. 
}

\textbf{Theorem. 3} \textit{
On any dataset pair that log-likelihood works well, \IE $\log \pin(x_1) > \log \pin(x_2)$ for most $x_1 \sim \pin, x_2 \sim \pout$, KL-based indicator can reach better performance.
}

\textbf{Theorem. 4} \textit{
For any likelihood-ratio indicator $\log \pin(x) - \log g(x)$ where $g$ is a continuous differentiable probability distribution, KL-based indicator outperforms them.
}

\textbf{Theorem. 5} \textit{
	$\log \frac{p_\theta(x)}{p_\gamma(x)}$ can reach better AUROC than KL-based indicator, when $p_\gamma$ is well-trained, \IE $p_\gamma$ reaches better likelihood on $\pmix$ than $p_{\hat{\gamma}} = \alpha p_{\theta} + \beta p_{\omega}$. }

By Theorem 1, indicator $\log p_\theta(x) - \log p_\omega(x)$ can nearly perfectly solve the OoD problem, as shown in Figure~\ref{fig2}. 
By Theorem~1, $\log \pin(x_{in}) - \log \pout(x_{in}) > 0$ for $x_{in} \in \pin$ and  $\log \pout(x_{out}) - \log \pin(x_{out}) > 0$ for $x_{out} \in \pout$, as shown in Figure~\ref{fig3}. 
$\log p_\theta(x) - \log p_\omega(x)$ is called \textbf{KL-based indicator}. However, $\log p_\theta(x) - \log p_\omega(x)$ is not a practical method since $\omega$ needs $D^{train}_{out}$, which is not allowed in OoD domain. Through Theorem~5, \textbf{\ROID{} find a method which does NOT need $\mathcal{D}^{train}_{out}$, to approximate KL-based indicator}, shown in the next section.

\begin{figure}
\center
\includegraphics[width=0.4\textwidth]{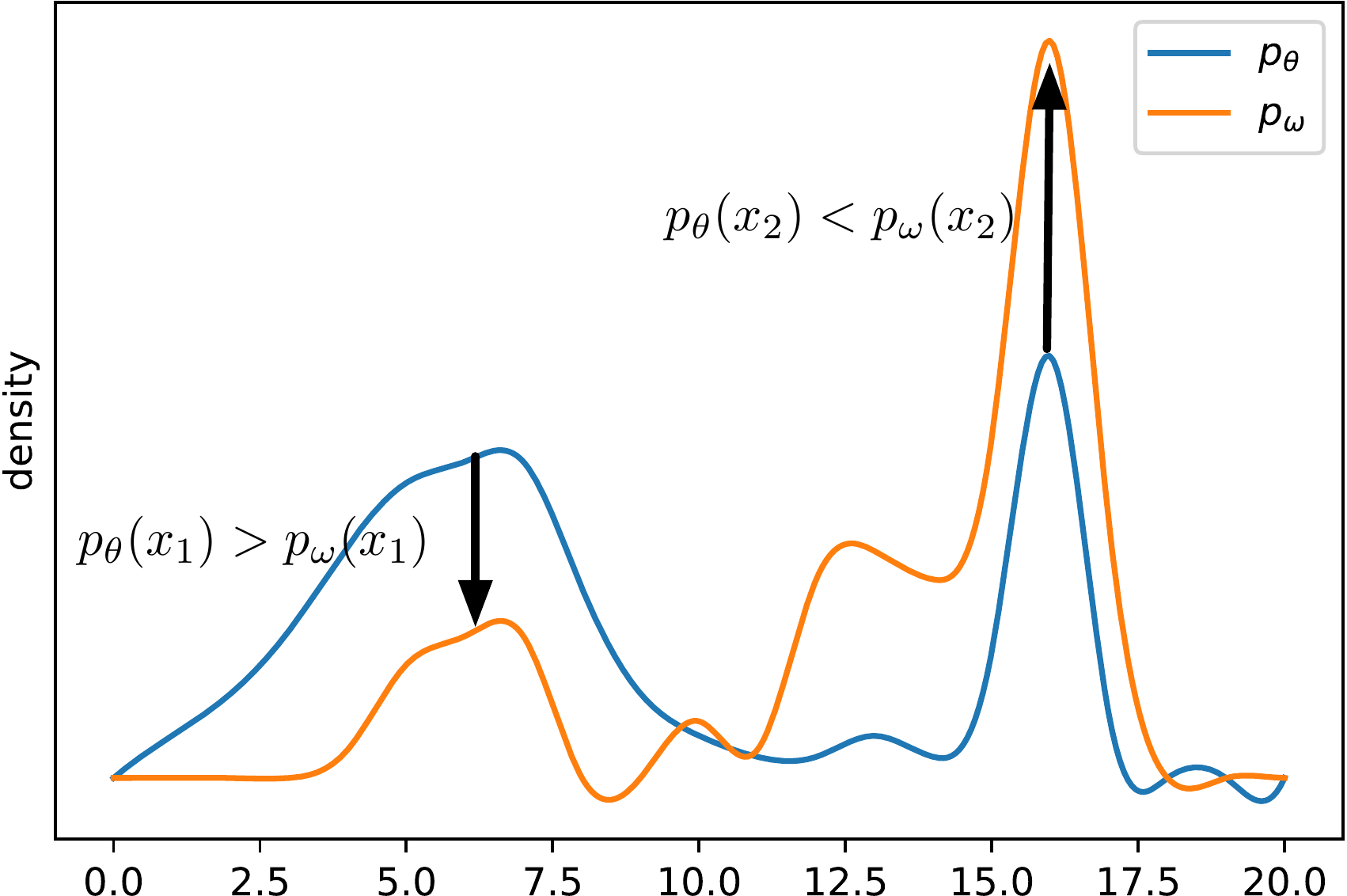}	
\caption{Diagrammatic sketch for KL-based indicator. The in-distribution is in [0, 10] and out-of-distribution in [10, 20]. Intuitively, $p_\theta$ assign higher density to $x_1 \in [0, 10]$ and lower
density for $x_2 \in [10, 20]$ than $p_\omega$, \IE $p_\theta(x_1) > p_\omega(x_1)$ and
$p_\theta(x_2) < p_\omega(x_2)$.}
\label{fig3}
\end{figure}

\subsection{Naive Fine-tune}
{
By Theorem~5, $\log p_\theta(x) - \log p_\gamma(x)$ could reach same performance as KL-based indicator, while $p_\gamma$ only needs $D^{test}$ for training (note that the training for $p_\gamma$ does NOT need the label in $D^{test}$).
From another perspective, using indicator $\log p_\theta(x) - \log p_\gamma(x)$ could  be treated as a fine-tune process: a model $p_\theta$ is well-trained on $D^{train}_{in}$, and then it is fine-tuned on $D^{test}$; \textbf{if $\bf{x \in D^{test}}$ gets a worse likelihood after fine-tuning, $x$ will be detected as in-distribution, otherwise out-of-distribution}.
Fine-tune criterion is significantly different from the previous likelihood criterion in Table~\ref{tab1}.

To introduce the fine-tune criterion, the \textbf{Naive Fine-tune} algorithm is proposed in Algorithm~\ref{alg:mixture}, which is only a strawman algorithm since Naive Fine-tune requires testing set $D^{test}$.
In Algorithm~\ref{alg:mixture}, $\frac{\beta}{\sigma} \|\log p_\theta(x) - \mu\|$ is used to detect whether the $p_\theta(x)$ is too high or too low ($x$ will be treated as OoD directly), by Theorem~1.
Fine-tune criterion is shown effective in Table~\ref{tab3}.

However, Naive Fine-tune will be useless in realistic scenes since the whole $D^{test}$ is hard to get~\cite{xu2018unsupervised}. The major weakness of Naive Fine-tune is that \textbf{Naive Fine-tune requires $D^{test}$. However, it is not allowed in the OoD domain}. To solve this problem, we did detailed researches about Naive Fine-tune in Section~\ref{subsec:concern} and developed an inductive method in the following subsection.

}

\begin{algorithm}[tb]
   \caption{Naive Fine-tune Algorithm}
   \label{alg:mixture}
\begin{algorithmic}
   \STATE {\bfseries Input:} The training set $\mathcal{D}^{train}_{in}$, the testing set $D^{test}$. $b$ represents whether use $\theta$ to initialize $\gamma$. $\beta = 0.1$. 
   \STATE {\bfseries Output:} Predicted label for each image in $D^{test}$
   \STATE Maximize log-likelihood $\log p_\theta$ on $\mathcal{D}^{train}_{in}$
   \STATE $\gamma \leftarrow \theta$ if $b$ is True else random initialize
   \STATE Maximize log-likelihood $\log p_\gamma$ on $\mathcal{D}^{test}$

   \STATE $\mu \leftarrow \E_{D^{train}_{in}} [ \log p_\theta(x)]$, $\sigma \leftarrow \mathop{Var}_{D^{train}_{in}} [\log p_\theta(x)]$
   \STATE \textbf{return} $\log \frac{p_\theta(x)}{p_\gamma(x)}- \frac{\beta}{\sigma} \|\log p_\theta(x) - \mu\| $ for each $x \in \mathcal{D}^{test}$
\end{algorithmic}
\end{algorithm}

\subsection{Single-shot Fine-tune}
{
By Theorem~5 and Figure~\ref{fig6}, Naive Fine-tune reaches a promised performance in few epochs.
Inspired by this, we propose Single-shot Fine-tune algorithm in Algorithm~\ref{alg:single}.
The key idea of Single-shot Fine-tune is \textbf{to fine-tune model on the single testing sample $x$, instead of the whole testing set}.
 Data-augmentation generates more samples to enhance the generality of $p_\gamma$.

Especially, if $x \in \pin$, $p_\gamma$ is fine-tuned on a sample from in-distribution (note $p_\theta(x)$ is well-trained on $\pin$) and then $\log p_\gamma(x)$ will be close to $\log p_\theta(x)$; if $x \in \pout$,  $p_\gamma$ is fine-tuned on a sample from out-of-distribution, and then $\log p_\gamma(x)$ will be much larger than $\log p_\theta(x)$.

Through fine-tuning model on the single testing sample, Single-shot Fine-tune solves the weakness of Naive Fine-tune. The input required by Single-shot Fine-tune is only the testing sample $x$, and obviously, every method needs $x$ as input. At last, algorithm~\ref{alg:single} might be confusing since it uses the back-propagation, which is usually used in training instead of testing.

\textbf{Why we can use back-propagation}. 
In canonical deep learning domains, \EG classification, back-propagation is usually used in training instead of testing because of the following 3 major reasons:

\noindent \textbf{Labels}. In the canonical deep learning domain, \EG classification, the loss function usually uses the labels, and then back-propagation needs labels as input, which is not allowed in testing. 

\noindent \textbf{Samples}. Back-propagation needs enough samples. However, in the testing stage, especially in the inductive learning domain, it is not allowed to obtain many testing samples.

\noindent \textbf{Time}. Back-propagation usually needs many steps to train the model, which is time-consuming. However, especially in some online-system, the testing time should be short enough.

The above problems lead to an inherent impression that back-propagation can not be used in the testing stage. 
However, Single-shot Fine-tune has solved the above problems in the OoD domain:

\noindent \textbf{Labels}. The loss functions of deep generative models do not use the labels.

\noindent \textbf{Samples}. Single-shot Fine-tune algorithm only uses the single testing sample $x$ as input and uses data-augmentation to enhance the generality of samples. Only popular data-augmentation methods are used (containing shift and rotation) instead of special-designed data-augmentation. 

\noindent \textbf{Time}. Single-shot Fine-tune only uses few steps (64 steps in 7s per testing sample, while Naive Fine-tune costs 60k steps) to fine-tune the model. The fastest method likelihood costs 0.36s per testing sample, but its performance is much lower (24\%) than Single-shot Fine-tune. 

In conclusion, we argue that back-propagation should be allowed to be used in Single-shot Fine-tune.


\begin{algorithm}[tb]
   \caption{Single-shot Fine-tune Algorithm}
   \label{alg:single}
\begin{algorithmic}
   \STATE {\bfseries Input:} The training set $\mathcal{D}^{train}_{in}$, the testing sample $x$, fine-tuning steps $m=64$, and $\beta=0.1$.
   \STATE {\bfseries Output:} Predicted labels for $x$
   \STATE Maximize log-likelihood $\log p_\theta$ on $\mathcal{D}^{train}_{in}$
   \STATE $\gamma \leftarrow \theta$
   \FOR{$i=1$ {\bfseries to} $m$}
   \STATE Generate a batch $b$ through data-augmentation for $x$
   \STATE Maximize log-likelihood $\log p_\gamma$ on $b$ for one step
   \ENDFOR
   \STATE $\mu \leftarrow \E_{D^{train}_{in}} [ \log p_\theta(x)]$, $\sigma \leftarrow \mathop{Var}_{D^{train}_{in}} [\log p_\theta(x)]$
   \STATE \textbf{return} $\log \frac{p_\theta(x)}{p_\gamma(x)} - \frac{\beta}{\sigma} \|\log p_\theta(x) - \mu\| $
\end{algorithmic}
\end{algorithm}

}
\section{Experiments}\label{sec6}
This section demonstrates the effectiveness of KL-based indicator, Naive Fine-tune, and Single-shot Fine-tune, on computer vision benchmark datasets. Detailed setup is shown in Appendix B. 

\subsection{Major Results}
\begin{table}[htbp]
\caption{The average AUROC of KL-based indicator (only theoretical, NOT practical), Naive Fine-tune (transductive), and Single-shot Fine-tune (inductive) on VAE, PixelCNN, and RNVP. The KL-based indicator uses $\mathcal{D}^{train}_{out}$. Thus it is not practical. Naive Fine-tune reaches nearly the same performance as the KL-based indicator, which validates the Theorem~5. Single-shot Fine-tune is slightly worse than Naive Fine-tune. However, it is inductive and only costs about 7s per image.}
\label{tab3}
\begin{center}
\begin{tabular}{lllllll}
Indicator     & VAE & PixelCNN & RNVP  \\
\toprule
KL-based Indicator & \textbf{99.08} & \textbf{99.85} & \textbf{99.81} \\
Naive Fine-tune & \textbf{98.68} & \textbf{97.80} & \textbf{98.55} \\
Single-shot Fine-tune & \textbf{95.78} & \textbf{97.64} & \textbf{94.34} \\
\bottomrule
\end{tabular}
\end{center}
\end{table}

The main results of past works are shown in Table~\ref{tab1}. The main results for KL-based indicator, Naive Fine-tune, and Single-shot Fine-tune are summarized in Table~\ref{tab3}.

\subsection{Addressing concerns}\label{subsec:concern}
The key idea of this paper is to detect OoD by fine-tuning the model with testing samples. However, there are the following major concerns about this idea:

\noindent \textbf{Q1.} Is Naive Fine-tune \textbf{data-specific}? \IE does it work for the data that have not been fine-tuned on?

\noindent \textbf{A1.} In Table~\ref{tab4}, Naive Fine-tune reaches promised performance (slightly lower than KL-based indicator) when 20\% testing data are used for fine-tuning. Thus Naive Fine-tune is not data-specific.

\noindent \textbf{Q2.} Can our method work \textbf{online}? \IE the testing data is streaming.

\noindent \textbf{A2.} We simulate an online system with streaming $\mathcal{D}^{test}$: $D^{test}=[x_1,\ldots,x_n]$, and, $x_1, \ldots, x_i$ is known when $x_i$ is testing. At time $i$, Naive Fine-tune runs with $[x_1, \ldots, x_i]$. Table~\ref{tab4} and Figure~\ref{fig6} shows that Naive Fine-tune reaches promised performance with online limitation.

\noindent \textbf{Q3.} Can Naive Fine-tune work if $\mathcal{D}^{test}$ contains \textbf{few data}?

\noindent \textbf{A3.} $\mathcal{D}^{test}$ is divided into several blocks, and Naive Fine-tune runs on each block. In this case, data for fine-tuning and fine-tune epochs are less than the ordinary case. Figure~\ref{fig6} shows that the optimization leads to unexpected $p_\gamma$ when data are insufficient. Training on past data (online) can alleviate the issue.
These experiments show that Naive Fine-tune is effective, simple, online, and not data-specific. It also shows the weakness of that Naive Fine-tune algorithm can not work well when data for Naive Fine-tune are severely insufficient. It encourages the development of Single-shot Fine-tune.

\noindent \textbf{Q4.} What is the difference between \textbf{pretrained} (initialize $\gamma$ with $\theta$) and \textbf{unpretrained} models?

\noindent \textbf{A4.} Figure~\ref{fig6} shows AUROC during Naive Fine-tune. Unpretrained model needs more epochs to reach a better performance than the pretrained model. In contrast, the pretrained model can easily reach a promised performance in few epochs, which leads to Single-shot Fine-tune algorithm.

{
\noindent \textbf{Q5.} Does the Single-shot Fine-tune method rely on the data augmentation and the number of steps?

\noindent \textbf{A5.} 
Algorithm 2 does not mandate a special-designed data-augmentation. 
Appendix B shows experiments with data augmentations containing
shifting, rotation, blur, noise, scale, cropping, flipping, and modifying contrast and lightness. Their performance has no significant difference ($\leq$0.3\%) to the basic data-augmentation containing rotation and shift. When step $< 64$, the step significantly influences the AUROC, and when step $> 64$, AUROC is nearly the same, as shown in appendix B.2. Therefore, we set step = 64 as the default parameter of Single-shot Fine-tune. Additionally, when step =64, the time-cost is only 7s per testing sample.
}

\begin{table}[htbp]
\caption{Average AUROC of Naive Fine-tune with limitations introduced in Section~\ref{subsec:concern}.
}
\label{tab4}
\begin{center}
\begin{tabular}{lllllll}
Limitation & None & online & 20\% & block \\
\toprule
 VAE & \textbf{98.68} & 97.24 & 96.50 & 97.60\\
PixelCNN  & \textbf{97.80} & 91.77 & 91.47 & 88.80 \\
RNVP  &  \textbf{98.55} & 90.92 & 88.16 & 90.44\\
\bottomrule
\end{tabular}
\end{center}
\end{table}

\begin{figure}[t]
\centering
\includegraphics[width=0.22\columnwidth]{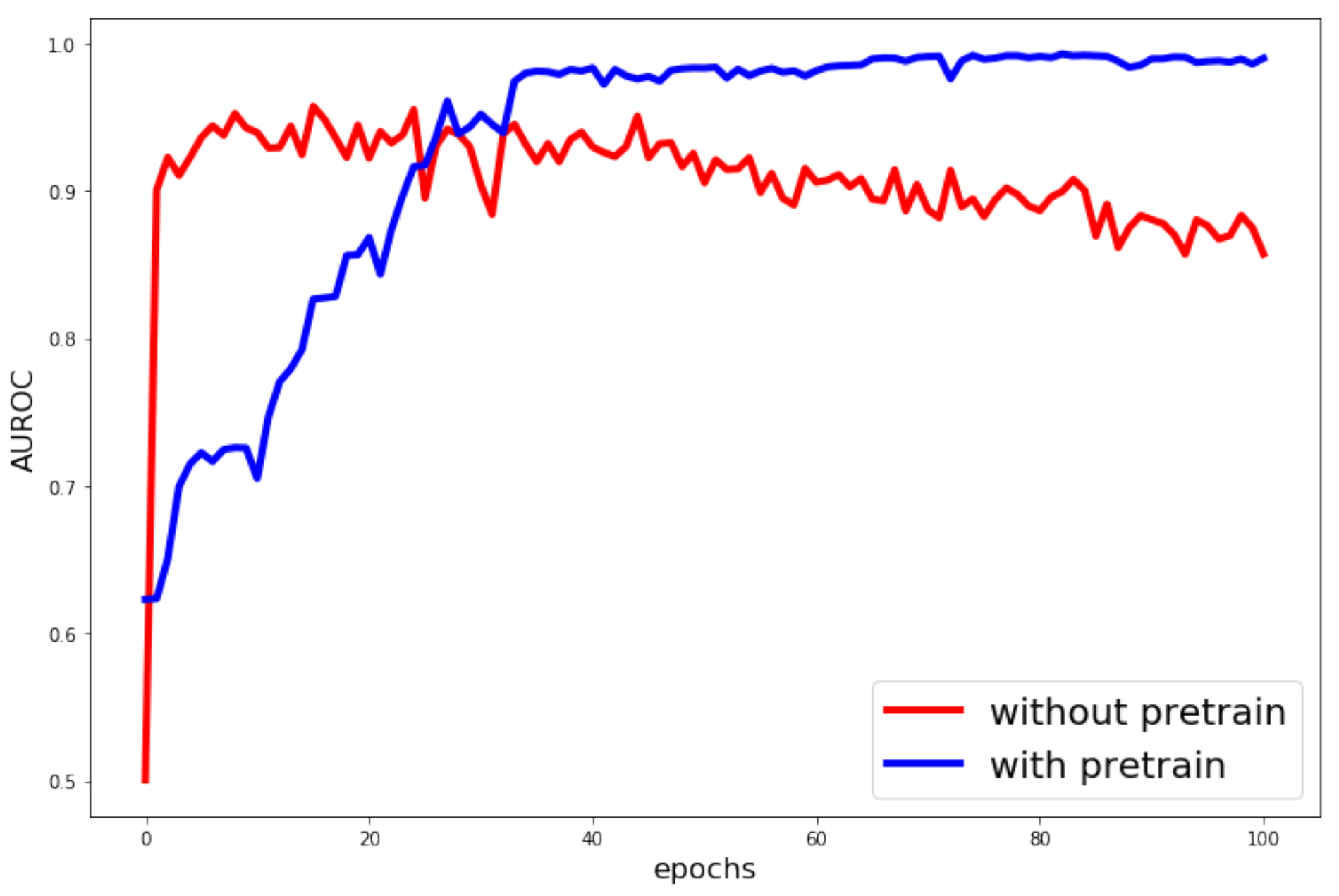}
\includegraphics[width=0.22\columnwidth]{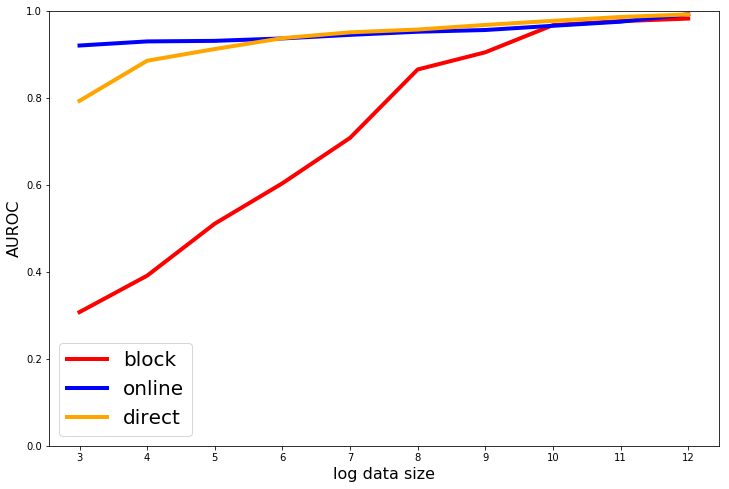}
\includegraphics[width=0.45\columnwidth]{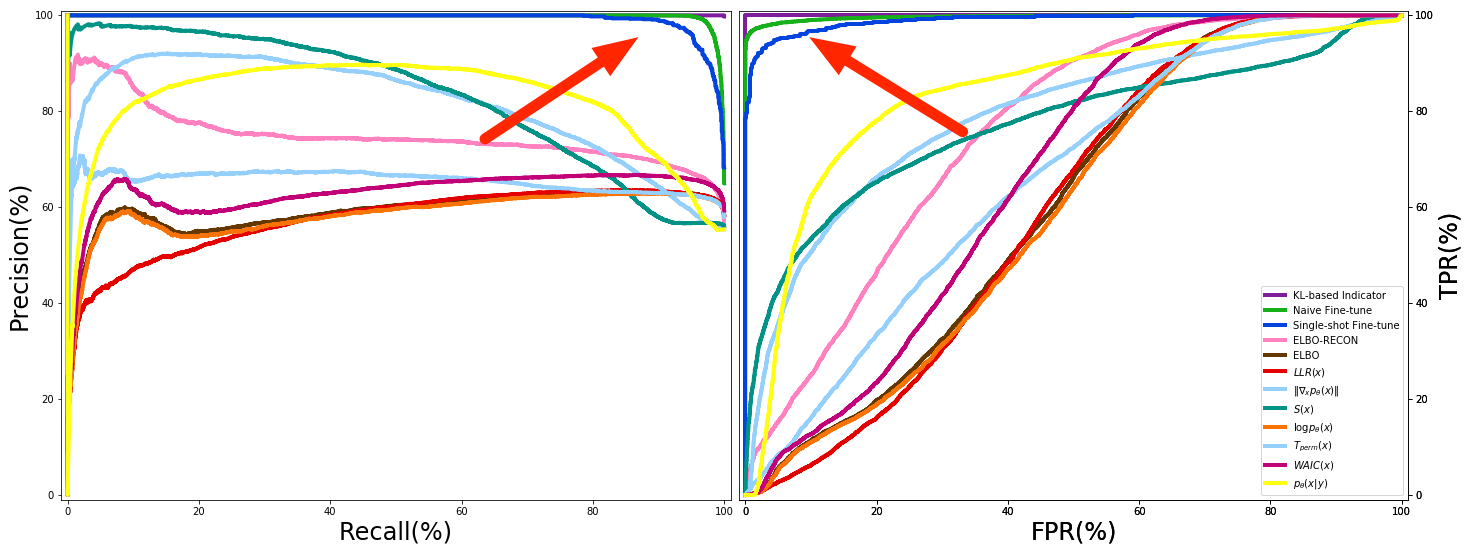}
\caption{Left: Average AUROC of pretrained model and unpretrained model during the Naive Fine-tune on CIFAR-10 vs other datasets. Mid: AUROC of Naive Fine-tune when the number of samples used by Naive Fine-tune varies on CIFAR-10 vs SVHN. 'online', 'block' is introduced in Section~\ref{subsec:concern}. 'direct' indicates that model $p_\gamma$ is directly trained on such few data with enough epochs. Right: ROC and PRC on MNIST vs Omniglot based on VAE model. The KL-based indicator, Naive Fine-tune algorithm, and Single-shot Fine-tune algorithm significantly outperform others. }
\label{fig6}
\end{figure}

\subsection{Validation of Theorem}

Theorem~1 and Theorem~3 are supported by the detailed experiments shown in appendix B. Theorem~2 is supported by Table~\ref{tab3}, where the performance of $\log p_\theta(x) - \log p_\gamma(x)$ is quite close to $\log p_\theta(x) - \log p_\omega(x)$. 
Figure~\ref{fig6} shows PRC and ROC of KL-based indicator, log-likelihood indicator, likelihood ratio indicator, and others, which supports Theorem~4 that KL-based indicator is the best. Theorem~5 is supported by Table~2, where the performance of Naive Fine-tune is close to the KL-based indicator. 


\subsection{Limitations of this Study}

\noindent \textbf{Limitation of datasets.}
In our paper, we use a large collection of benchmarks to show the generality of indicators. However, we only consider natural OoD datasets and do not consider attacked OoD, which are categorized by \cite{chen2020robust}. An important reason is that there is no universal criterion like simply-classified, to measure attacked OoD datasets.

\noindent \textbf{Limitation of models.}
In our paper, for fair comparison and generality, we only consider the common models, including ResNet, VAE, PixelCNN, RealNVP, and WGAN. However, there are numerous models careful-designed for OoD detection. Due to the resource limitation, we can not provide the performance of them on the large collection of benchmarks.

\noindent \textbf{Limitation of KL-based indicators.}
In section~\ref{sec6}, KL-based indicators rely on the model (Naive Fine-tune with PixelCNN and RNVP is more data-specific) and optimizer (optimizer can not provide the expected $p_\gamma$ with insufficient data).
Single-shot Fine-tune solves such problems through data augmentation. However, Single-shot Fine-tune reaches worse performance than Naive Fine-tune, and it needs to be developed (\EG careful-designed data-augmentation and optimizer for single-shot).

\section{Conclusion and Future Work}
This paper first shows none of the existing OoD indicators based on deep generative models perform well on the large collection of benchmarks. We then propose a novel theoretical framework \ROID{} for divergence-based out-of-distribution indicators and propose the Single-shot Fine-tune algorithm, which significantly outperforms past works by 5$\sim$8\% in AUROC.

We believe the divergence-based out-of-distribution indicator theoretical framework and fine-tune criterion of our paper are important steps towards developing more effective OoD indicators based on deep generative models. For future work, it will be interesting to propose more OoD indicators through \ROID{} framework and fine-tune criterion.
 
\newpage
\bibliography{neurips_2021}
\bibliographystyle{plain}

\end{document}